\documentclass[10pt,letter,conference]{IEEEtran}

\usepackage{times}
\usepackage{epsfig}
\usepackage{graphicx}
\usepackage[usenames,dvipsnames]{xcolor}
\usepackage{amsmath}
\usepackage{amssymb}
\usepackage[hyphens]{url}
\usepackage{hyperref}
\hypersetup{
    breaklinks=true,
    urlbordercolor=White
}

\usepackage{cite}
\usepackage{float}
\usepackage{amsmath,amssymb,amsfonts}
\usepackage{graphicx}
\usepackage{textcomp}
\usepackage{multirow}
\usepackage{caption}
\usepackage{subcaption}
\usepackage[margin=1.0in]{geometry}
\usepackage{algorithm}
\usepackage{algorithmicx}
\usepackage{algpseudocode}

\makeatletter
\renewcommand{\ALG@beginalgorithmic}{\scriptsize}
\makeatother

\begin{document}

\title{Examining Uniqueness and Permanence of the WAY\_EEG\_GAL dataset toward User Authentication 
}

\author{
\IEEEauthorblockN{Aratrika Ray-Dowling}
\IEEEauthorblockA{
\textit{ray@clarkson.edu}\\
\textit{Department of Electrical and Computer Engineering, Clarkson University}\\
\textit{8 Clarkson Avenue, Potsdam, New York 13699, USA}
}
}

\maketitle

\begin{abstract}
Physiological biometric based security mechanisms have been deployed in different areas like border control, secured access points, national identification, smart devices, online banking, and others.
The most popular physiological biometric modalities utilized for such purposes are face, iris, fingerprint, and palm print.
There are notable state of the art that have evaluated the usability of Electroencephalography (EEG) as a physiological biometric modality for security and authentication purposes.  
This study evaluates the discriminating capacity (uniqueness) of the EEG data from the WAY\_EEG\_GAL~\cite{6-Luciw2014} public dataset to authenticate individuals against one another as well as its permanence.
In addition to the EEG data, Luciw et al.~\cite{6-Luciw2014} provide EMG (Electromyography), and kinematics data for engineers and researchers to utilize WAY\_EEG\_GAL~\cite{6-Luciw2014} for further studies. 
However, evaluating the EMG and kinematics data is outside the scope of this study. 
The goal of the state-of-the-art~\cite{6-Luciw2014} is to determine whether EEG data can be utilized to control prosthetic devices. 
On the other hand, this study aims to evaluate the separability of individuals through EEG data to perform user authentication. 
A feature importance algorithm is utilized to select the best features for each user to authenticate them against all others. 
The authentication platform implemented for this study is based on Machine Learning models/classifiers.
As an initial test, two pilot studies are performed using Linear Discriminant Analysis (LDA) and Support Vector Machine (SVM) to observe the learning trends of the models by multi-labeling the EEG dataset. 
As one of the controlled variables in this Machine Learning pipeline
kNN, LDA, and SVM, are picked as classifiers for both pilot tests and user authentication as in the recent studies~\cite{abdulrahman2020support,1-Bashar2016,chin2019exploring,5-Jayarathne2020,Nakamura2017_ear} it is observed that the usage of these classifiers produces high authentication performance. 
Utilizing kNN first as the classifier for user authentication, accuracy around $\approx 75$\% is observed. 
Thereafter to improve the performance both linear and non-linear SVMs are used to perform classification. The overall average accuracies of 85.18\% and 86.92\% are achieved using linear and non-linear SVMs respectively. In addition to accuracy, F1 scores are also calculated. 
The overall average F1 score of 87.51\% and 88.94\% are achieved for linear and non-linear SVMs respectively. Beyond the overall performance, 
high performing individuals with 95.3\% accuracy (95.3\% F1 score) using linear SVM and  97.4\% accuracy (97.3\% F1 score) using non-linear SVM are also observed.

\end{abstract}
\vspace{1em}

\section{INTRODUCTION}
\label{sec:intro}

User authentication utilizing Electroencephalography (EEG) modality is feasible provided it possesses the following biometric characteristics, namely, universality, uniqueness, permanence, and collectibility~\cite{Nakamura2017_ear}.  
In contrast to other established physiological biometrics (face, iris, fingerprint, palm), EEG has anti-spoofing characteristic as it is difficult to forge complex signals from several EEG channels of a user profile. Whereas, both active (sophisticated effort) and passive (unsophisticated effort) types of spoof attacks can be performed on face, fingerprint, iris, and palm biometrics. 
However, the biometric characteristics of acceptance (in terms of user friendliness) and  permanence need to be enhanced to deploy EEG based authentication systems in real life. Hence, studies on enhancing acceptance and permanence of EEG data for authentication are always encouraged. 

Most commonly, EEG data is acquired using wearable devices like Acti-Cap or electrode headsets which requires a degree of initial preparation like placing several electrodes in addition to gel application on the user's scalp. This may throw some hindrances in the process of the  data collection. 
Although, the user does not have to interact with any interface to provide the data. 
To make user friendly EEG data acquisition Kosmyna et al.~\cite{Kosmyna2019attentivu_glass} have
 developed a wearable pair of EEG and EOG (Electrooculography) glasses. In the work by Zhang et al.~\cite{zhang2019identity} a portable device, called MindWave Mobile, is used for EEG data acquisition via user's forehead. The single-channel device simplifies the initial preparation process and improves user comfort. The data collected through this portable device has authenticated users with accuracies ranging between 80\% and 95\%. Further developments of such EEG devices with fewer electrodes, which can be made commercially available, can enhance the acceptance/user friendliness characteristic of EEG based biometrics system.

Depending on the type of security application, the EEG based user authentication system may perform template-based user enrollment, where a user needs to submit several data samples to build templates for future use~\cite{abuhamad2020sensor}. Authenticating users in a future time frame requires EEG to have permanence characteristics. This can be introduced by keeping the user behavior/activity constant during both enrollment and verification. Here, the conditions of the surrounding environment must also be considered. User authentication through EEG data can be performed by three data acquisition processes - (i) while relaxing with eyes closed; (ii) while exposed to visual stimuli; and (iii) while performing mental tasks~\cite{1-Bashar2016}.

 In this study, the EEG data is utilized from  the WAY\_EEG\_GAL~\cite{6-Luciw2014} public dataset, which is acquired while users are performing a mental task of lifting, holding, and resting objects of different weights (165g/330g/660g) and surfaces (sandpaper/suede/silk) where both parameters (weight and surface texture) get changed randomly. 
 In the process, each user provides several grasp and lift trials for at least two hours. Successfully authenticating users across this time frame of two hours during which each user's behavior/activity is similar
 will ensure the permanence characteristic of
 the WAY\_EEG\_GAL~\cite{6-Luciw2014} EEG data to a considerable degree. Therefore, the aim of this study lies in evaluating the user discriminative capacity 
 of the EEG data, acquired across the span of two hours while users performed the mental task of lifting objects. Figures~\ref{fig:P1_raw} and~\ref{fig:P12_raw} show user 1 and 12's time domain data respectively where we can visually see a difference in the data trend between session 1 and session 9. Acceptable authentication accuracy will ensure the permanence of this dataset given its dynamic nature across sessions.  
 
 The state of the art public dataset~\cite{6-Luciw2014} is aimed for acquisition of EEG signals for prosthetic control of object manipulation. Through this, one can study the precision grasp-and-lift (GAL) of an object. The dataset has 32 channels of EEG data, 5 channels of EMG data, and kinematics data. 
 However, in this study, the EMG and kinematics data are not utilized. 
 Here, user authentication is performed only through the EEG data to evaluate its degree of uniqueness and permanence, while users are executing grasp-and-lift trials of objects.  

After the initial pre-processing, statistical features are extracted over each of the 32 channels of the EEG data which increases the data dimensionality.
Therefore, feature importance is performed per genuine user that chooses the best features to authenticate each genuine user against the rest.  

State of the art on EEG based user authentication show high performances when Machine Learning classifiers, namely, kNN (~\cite{jayarathne2017-intro},~\cite{2-Rahman2021},~\cite{Valizadeh2019_decrypting}), SVM   (~\cite{4-Albasri2019},~\cite{1-Bashar2016},~\cite{2-Rahman2021},~\cite{Valizadeh2019_decrypting},~\cite{koike2016high}), and LDA (~\cite{chin2019exploring},~\cite{5-Jayarathne2020}) are utilized. 
Hence, in this study, for the two pilot tests and user authentication experiments,   
kNN, SVM, and LDA classifiers are chosen. 
The two pilot studies performed utilize LDA and non-linear SVM (Radial Basis Function) which achieve accuracies of 53.3\% and 68\% respectively on multi-labeled data (each user had a unique label). 
The pilot studies give an idea of the learning trends of these Machine Learning models on the multi-labeled  WAY\_EEG\_GAL data. 
Observing the pilot tests, the experiments on user authentication are performed where each genuine user is authenticated against all other impostor users. 
For this, initially, kNN classifier is used, which achieves around $\approx 75$\% accuracy. 
Looking at the pilot tests and the first attempt of user authentication using kNN, finally SVM (both linear and non-linear) models are chosen to perform classification. 
The overall average accuracies of 85.18\% and 86.92\% across users are achieved using linear and non-linear SVMs respectively. 
Also, across all users, the overall average F1 scores of 87.51\% and 88.94\% are observed using linear and non-linear SVMs respectively. Beyond the overall performances, individuals showing high performances of 95.3\% accuracy (95.3\% F1 score) with linear SVM and  97.4\% accuracy (97.3\% F1 score) with non-linear SVM are also observed.

Therefore, this work has the following contributions:

i) Utilizing WAY\_EEG\_GAL public dataset~\cite{6-Luciw2014} for user authentication.

ii) Evaluating the discriminating (uniqueness) and permanence characteristics of the EEG data from the public dataset~\cite{6-Luciw2014}.

iii) Performing feature importance per genuine user for authenticating against all other impostors. 

iv) Utilizing both Linear and non-linear SVMs for user authentication on this dataset. 

v) A thorough statistical test to evaluate the significant difference of the performance trends across all users between linear and non-linear SVMs. We observe a close performance between the linear and non-linear SVMs.

The rest of the work includes Section~\ref{sec:lit_rev} which presents the related works. 
Section~\ref{sec:data} describes the public dataset. 
In Section~\ref{sec:method} we discuss the experimental procedures. Section~\ref{sec:results} reports all the experimental results.
Section~\ref{sec:hypo_test} reports hypothesis testing.
Lastly, Section~\ref{sec:conclusion} concludes our study.

\section{RELATED WORKS}
\label{sec:lit_rev}

The criteria of choosing the state of the art studies
that are related to this work is based on the Machine
Learning algorithms utilized for user authentication and
the user behavior involved during acquisition of the
data. Most recent studies tend to expose users to a
stimuli (visual or auditory) or keep them in resting state
while the EEG data is acquired. 
Given the data collection procedure in WAY\_EEG\_GAL involves lifting objects of random weights and textures, authenticating users from this dataset is based on individuals performing mental tasks for 2 hours. 
The aim of working on this dataset~\cite{6-Luciw2014} for user authentication through EEG is to evaluate the uniqueness and permanence  of the data collected while users are engaged in performing mental tasks of lifting random objects. 
The test of uniqueness will indicate the discriminating capacity of this dataset for user authentication through mental tasks. 
On the other hand, the test of permanence of this data will 
estimate the authentication capcity of EEG data over a time span
of 2 hours given EEG is dynamic compared to most
biometrics.

The study by Bashar et al.~\cite{1-Bashar2016} claim the validity of utilizing EEG as a biometric modality. Individual's brain signals are linked to their genetic information which make them unique from other individuals. Collecting data from 9 subjects through EMOTIV headset, they classify individuals using multi-class SVM classifier. The data is collected at a sampling rate of 128 Hz from 5 EEG channels (AF3, AF4, T7, T8, Pz) per user for six weeks. Over the span of several weeks of data collection they hold several sessions where each involve two tasks of eyes close and eyes open in relaxing state with minimal body movements.
They achieve best performances of 94.44\% of TPR (True Positive Rate) and 5.56\% of FPR (False Positive Rate). 

Nakamura et al.~\cite{Nakamura2017_ear} perform a feasibility test of the in-ear EEG sensor in their study. The in-ear sensor earpiece, consisting of two electrodes, is inserted in the user's left ear canal. It is made of a memory-foam substrate and two conductive flexible electrodes where the material is a viscoelastic foam which makes it a generic earpiece. Such devices enhance the user-friendliness of EEG based authentication once deployed in real life. 
Data is collected at a sampling rate of 1200 Hz from 15 subjects for two days. Users provide data in eyes close and resting state.
Utilizing binary SVM they achieve 99\% accuracy as the best result. 

In a recent work by Rahman et al.~\cite{2-Rahman2021} EEG data from 10 users is fused with their keypress data. Each of the 10 volunteers participate for 10 sessions where in each session a user types "qu-ELEC371" fixed-text password for 50 times. 
EEG has anti-spoofing capacity but may lack accuracy due to variability~\cite{2-Rahman2021}. This shows the need for test of permanence of EEG as a biometric modality from the research community. 
The EEG data is collected using EMOTIV headset (from AF3, AF4, T7, T8, PZ channels) while users are performing the mental task of typing the password.
The EEG data undergoes pre-processing using baseline correction, filtering, segmentation, and resampling techniques.
They extract features from the EEG data which includes statistical, frequency domain, and time domain (including the notable Hjorth parameters) features. 
They achieve 99.7\% and 98.7\% accuracies using only keypress and EEG respectively. They report the enhanced individual user performance when both modalities are fused at score level.

The study by Yang et al.~\cite{yang2019improved} involves evaluation of two public datasets, namely, UCI (Univeristy of California Irvine) EEG dataset~\cite{UCI_data} and EEG MMI (Motor Movement/Imagery) Dataset~\cite{MMI_data} for user authentication. 
The UCI~\cite{UCI_data} and MMI~\cite{MMI_data} datasets involve 122 and 109 volunteers respectively. In each dataset the EEG data is collected from EEG headsets with 64 channels. 
The UCI~\cite{UCI_data} and MMI~\cite{MMI_data} EEG data are collected at the sampling rates of 256 Hz and 160 Hz respectively. 
The data collections involve activities like eyes open, eyes close, and imagery tasks. Utilizing LDA (Linear Discriminant Analysis) classifier, Yang et al.~\cite{yang2019improved} achieve best accuracies of 93.28\% and 98.24\% with UCI~\cite{UCI_data} and MMI~\cite{MMI_data} datasets respectively.

The work by Arnau-Gonz{\'a}lez et al.~\cite{Arnau2018_influence} is another study that uses public datasets for EEG based user authentication. 
They utilize DEAP (Dataset for Emotion Analysis using EEG, Physiological and Video Signals)~\cite{DEAP_data}, MAHNOB-HCI~\cite{MAHNOB_data}, and SEED (SJTU Emotion EEG (SEED) Dataset)~\cite{SEED_data} datasets for user authentication that involve 32, 30, and 15 volunteers respectively. The EEG data of DEAP~\cite{DEAP_data}, MAHNOB-HCI~\cite{MAHNOB_data}, and SEED~\cite{SEED_data} are collected at sampling rates of 512 Hz, 512 Hz, and 1000 Hz respectively which are downsampled in Arnau-Gonz{\'a}lez et al.'s~\cite{Arnau2018_influence} study.
Each dataset exposes volunteers to a video stimuli to capture variations in emotion. 
Involving several machine learning and deep learning algorithms, this work achieves a best result of 99.15\% accuracy with the DEAP dataset~\cite{DEAP_data}. 

In this work, we utilize WAY\_EEG\_GAL public dataset~\cite{6-Luciw2014} and perform Machine Learning based classification for user authentication. We have performed feature importance per user profile using ExtraTreeClassifier algorithm. Both linear and non-linear SVMs are utilized for classification so that the difference in the performances can be observed between the two algorithms belonging to the same family. 

\begin{figure*}[t]
         \centering
         \fbox{\includegraphics[width=0.85\linewidth]{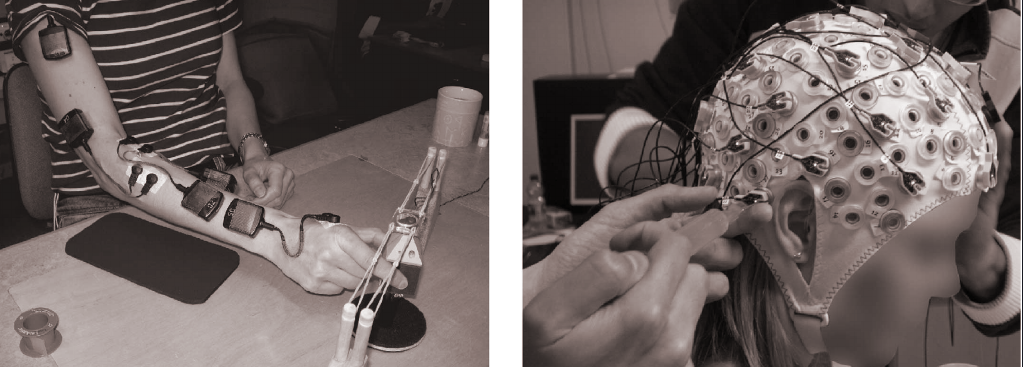}}
         \caption{\small Data collection set up of WAY\_EEG\_GAL~\cite{6-Luciw2014}. See Appendix~\ref{FirstAppendix} for EEG electrode labelings.}
         \label{fig:data_collect}
\end{figure*}

\begin{table*}[h]
\centering
\small

\begin{tabular}{ |c|c|c|c|c|c|c| } 
\hline
                                  
\textbf{Data status}            &TOTAL  &AVERAGE &MEDIAN    &MINIMUM     &MAXIMUM  &STANDARD \\ 
                        &&&&&                                                    &DEVIATION  \\\hline
                        
\textbf{Pre-processing}   &19548778  &1629064.83 &1632897.5 &1447770 &1858867 &125591.34     \\\hline
\textbf{Post-processing}  &310137 &25844.75 &25905 &22966 &29493 &1993.59    \\\hline

\end{tabular}
\caption{\small Data statistics for all 9 sessions across 12 users in pre-processing and post-processing stages.}
\label{table:data_stat_pre_post}
\end{table*}

\begin{figure*}[t]
         \centering
         \fbox{\includegraphics[width=0.9\linewidth]{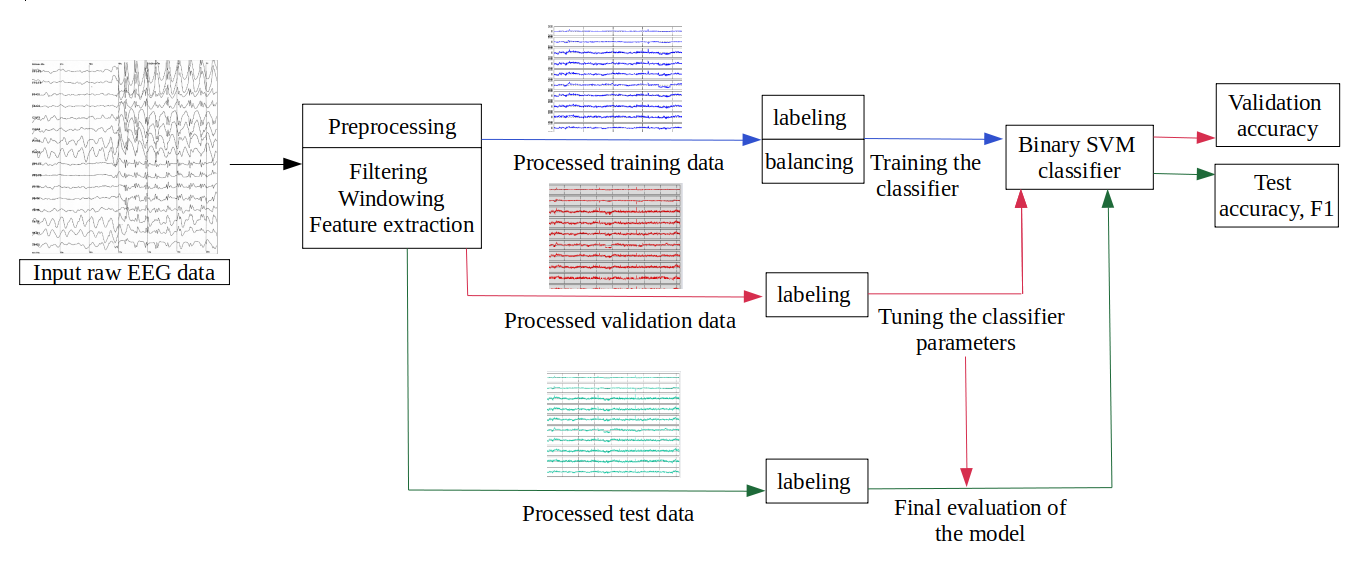}}
         \caption{\small The Machine Learning pipeline utilized for this study.}
         \label{fig:pipeline}
\end{figure*}

\section{WAY\_EEG\_GAL DATASET~\cite{6-Luciw2014}}
\label{sec:data}

The WAY\_EEG\_GAL~\cite{6-Luciw2014} (WAY : Wearable interfaces for hAnd function recoverY; EEG : Electroencephalography; GAL : Grasp And Lift trials) dataset has 12 volunteers (8 females, 4 males; 19-35 age group; right-handed) where each user participates for at least two hours providing around 328 grasp-and-lift trials. 
In total there are 10 sessions of data per user out of which it is observed that the $10^{th}$ session does not have corresponding EMG data.
Hence, to maintain uniformity we work with the first 9 sessions of each user. 
Table~\ref{table:data_stat_pre_post} shows the data statistics in post-processing stage across the 12 users for all the 9 sessions.
Each user is prompted to grab, lift, hold, and rest objects of different weights (165g, 330g, 660g) and surfaces (sandpaper, suede, silk surface) which are changed randomly during the data collection process during which the EEG data (along with other data) is collected. 
The random changes of object's weight and surface also 
contribute to the dynamic nature of the EEG dataset. 
The entire user activity (mental task) involves prompted user reaching out for the object, grasping it with thumb and index finger, lifting and holding it for a couple seconds, and finally putting it back on the support surface, thereafter releasing it and returning the hand to a designated rest position. 
The EEG is captured using a 32-channel Acti-cap where the sampling rate per channel is 500 Hz. Figure~\ref{fig:data_collect} shows the EEG data collection set up while user performs GAL trials.
The original target of the state of the art~\cite{6-Luciw2014} is to utilize EEG signals for prosthetic control of object manipulation. The EEG scalp recording while GAL trials decodes sensation, intention, and action of each individual.

\section{EXPERIMENTAL PROCEDURES}
\label{sec:method}
This section describes the Machine Learning pipeline followed to implement user identification. See Figure~\ref{fig:pipeline}.

\subsection{Pilot tests}

\begin{figure}[t]
         \centering
         \fbox{\includegraphics[width=0.75\linewidth]{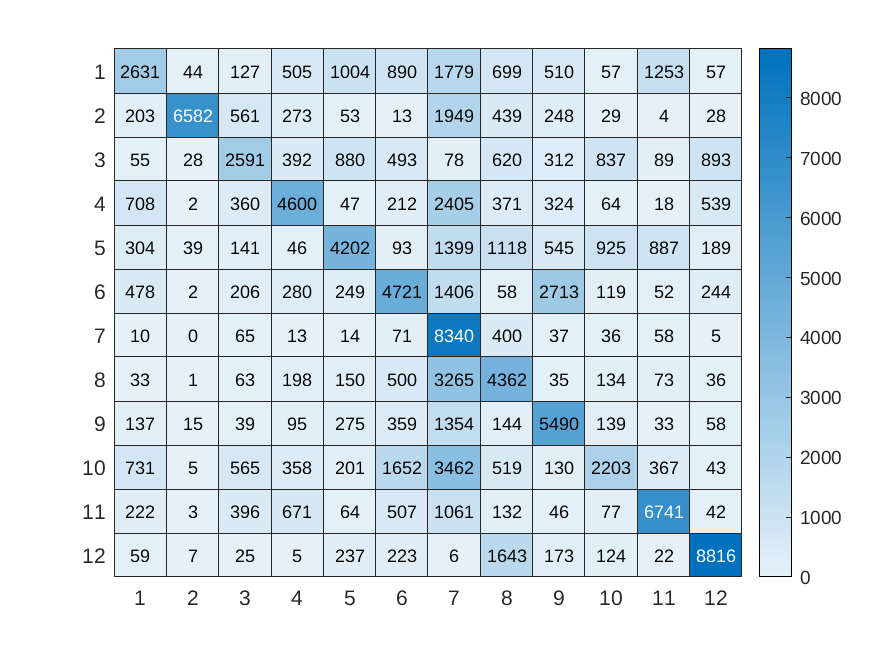}}
         \caption{\small Confusion matrix of pilot testing using Linear Discriminant Analysis.}
         \label{fig:pilot_RF}
\end{figure}

\begin{figure}[t]
         \centering
         \fbox{\includegraphics[width=0.75\linewidth]{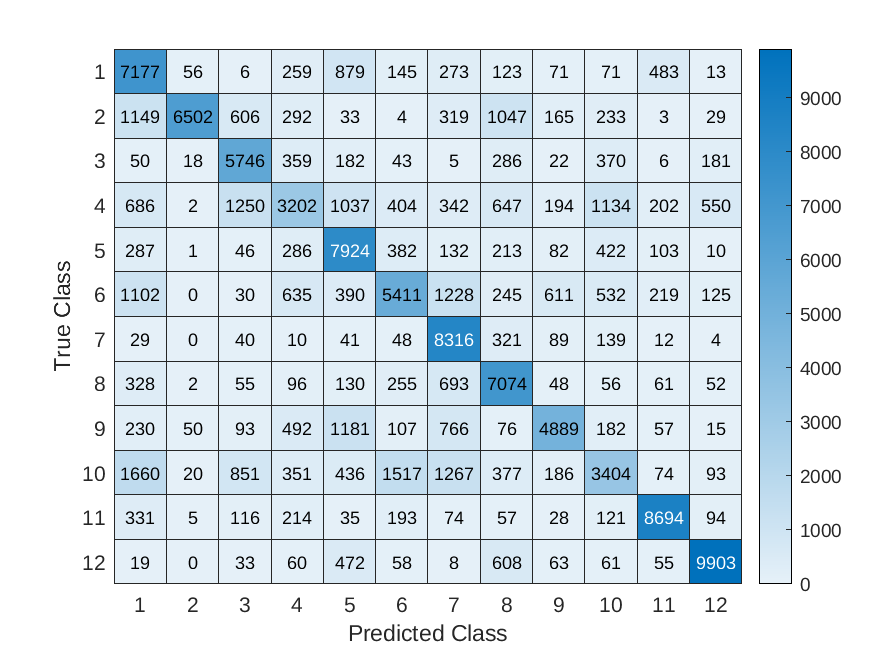}}
         \caption{\small Confusion matrix of pilot testing using Support Vector Machine.}
         \label{fig:pilot_SVM}
\end{figure}

Two pilot tests are performed using LDA (Linear Discriminant Analysis) and SVM (Support Vector Machine) keeping an 80:20 data split for training:testing.  
In the pilot tests, 12 labels for twelve users' data are used and a multi-class classification is performed.
This is to observe the learning capacities of the models on the dataset~\cite{6-Luciw2014}. The LDA with an eigensolver produces an accuracy of 53.3\% and the SVM with C=0.1 and gamma=scale as parameters produce an accuracy of 68\%. 
Actual user authentication experiments of genuine user versus impostors are performed after observing the results of true positives, true negatives, false positives, and false negatives from the confusion matrices in figures~\ref{fig:pilot_RF} and~\ref{fig:pilot_SVM}. The procedures of the actual authentication experiments are explained below.

\subsection{Data Preprocessing}

The sampling rate per channel of the EEG data is 500 Hz. 
A bandpass filter between 0.2 Hz and 45 Hz is used to obtain the EEG signal. 
All original 32-channel EEG data are retained and no pruning is performed.
All the 9 sessions from each user's data are taken where corresponding EMG data is available. The EMG data ensures the performance of GAL trials by users in all 9 sessions, although using the EMG data is out of the scope of this authentication study. 
See Appendix~\ref{SecondAppendix} for raw data visualization.
To each user's data, a window of 250 milliseconds with a 50\% overlap is applied. Hence, in every second 8 windows are available $\approx 125$ samples in each. 

\begin{figure}[h]
         \centering
         \fbox{\includegraphics[width=0.9\linewidth]{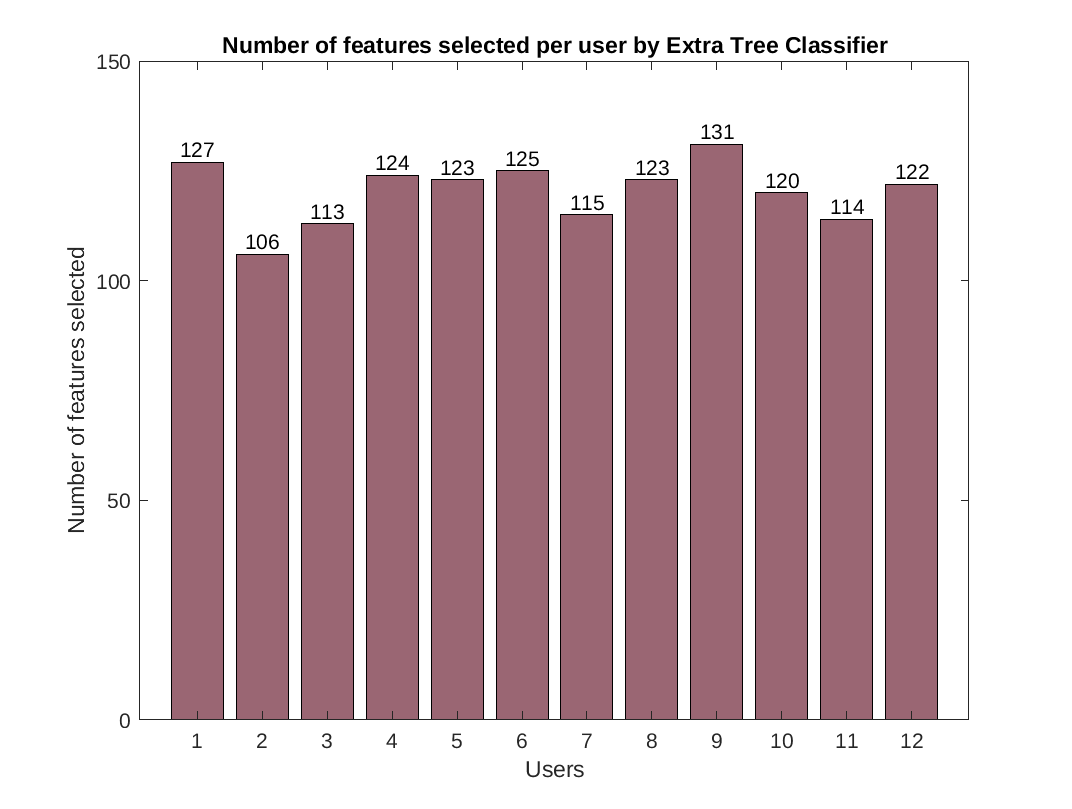}}
         \caption{\small The number of important features chosen per user by the ExtraTreeClassifier algorithm.}
         \label{fig:feat_select}
\end{figure}

\subsection{Feature extraction and feature selection} 

Univariate statistical features - average, standard deviation, root mean square, mean absolute value, skewness, and kurtosis are extracted from each of the 32 EEG channels per window. 
The simplified statistical features can reduce data dimensionality, reduce signal-to-noise ratio, and enhance classifier performance~\cite{gupta2020next}. 
Therefore, in total there are $32 channels * 6 features = 192$ feature columns. 
The features are normalized after extraction.
Features from sample users are plotted under Appendix~\ref{ThirdAppendix}. 
Given the feature dimension, a threshold-based feature selection algorithm (ExtraTreeClassifier) is used. ExtraTreeClassifier feature selection algorithm calculates feature importances.
The pre-tuned threshold decides the number of features that are significant to a user.
The number of important features selected per user by the ExtraTreeClassifier is shown in Figure~\ref{fig:feat_select}. 
The algorithm for ExtraTreeClassifier is shown below.

\begin{algorithm}[h]
    \caption{ExtraTreeClassifier Feature Importance}
    \label{alg:feat_imp}
    \begin{algorithmic}[1]
       \State df $\gets$ training feature columns
       \State model $\gets$ ExtraTreesClassifier()
        \State model.fit(df, training labels)
        \State feature\_importance $\gets$ sorted(model.feature\_importances\_)
        \State imp\_feat $\gets$ [] 
        \Comment\textit{array to store indices of important features}
        \For {$i \in$ range(len(feature\_importance))}
            \If {feature\_importance[i] $>=$ threshold}
                \State imp\_feat.append(i)
            \EndIf
        \EndFor
    \end{algorithmic}
\end{algorithm}

Appendix~\ref{FourthAppendix} shows a sample visualization of important features of a user with tuned user-specific threshold.

Table~\ref{table:data_stat_pre_post} shows the data statistics under post-processing stage which is the statistics of the data after it is processed to this stage.

\subsection{Data Splitting}

\begin{figure}[t]
         \centering
         \includegraphics[width=0.85\linewidth]{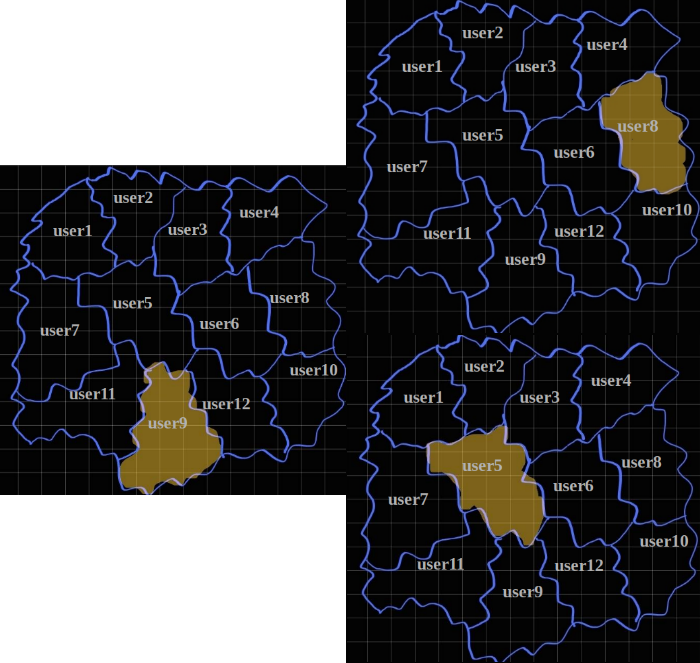}
         \caption{\small Schematic diagram showing one versus many authentication scenario.}
         \label{fig:one_vs_many}
\end{figure}

The EEG data is split into non-overlapping train, validation, and test sets. 
For each user, the first 5 sessions out of the 9 sessions are used for training, the next 2 sessions are used for validation, and the remaining 2 are used for testing. After processing the data up to feature normalization, the number of training, validation, and testing samples of all users are 195290, 65038, and 49809 respectively. The validation and testing samples differ by $\approx 20000$ because the duration of the sessions varies for users. We label the three sets of data with 1 for genuine users and 0 for all other users (impostors). Each genuine user is authenticated against all other impostors. Figure~\ref{fig:one_vs_many} shows a schematic diagram of a one versus many authentication scenario where the orange highlighted user to be authenticated is genuine and all others are impostor users.  
The training data is balanced by downsampling each impostor's data such that the sum of the total impostor samples closely matches to the total number of genuine samples.

\begin{table*}[h]
\centering
\small

\begin{tabular}{ |c|c|c||c|c|c|c| } 
\hline
                                  
\textbf{Performance}   &\textbf{VAL ACC}  &\textbf{VAL ACC} &\textbf{TEST ACC}    &\textbf{TEST ACC}     &\textbf{TEST F1 score}  &\textbf{TEST F1 socre}  \\
 \textbf{Statistics}                     &LSVM           &NLSVM      & LSVM       &NLSVM    &LSVM     &NLSVM \\\hline
\textbf{Average}      &86.37\%              &88.81\%             &85.18\%           &86.92\%           &87.51\%        &88.94\%       \\\hline
\textbf{Maximum}      &95.40\%              &98.40\%             &95.30\%           &97.40\%           &95.30\%        &97.30\%     \\\hline
\textbf{Minimum}      &79.80\%              &76.40\%             &76.40\%           &78.90\%           &81.00\%        &83.20\%   \\\hline
\textbf{Median}       &85.30\%               &88.75\%             &83.95\%           &84.85\%           &86.50\%        &87.25\%   \\\hline
\textbf{Std. Dev.}    &4.91\%               &5.78\%              &5.48\%            &6.43\%            &4.08\%         &4.94\%   \\\hline

\end{tabular}
\caption{\small Overall performance of linear SVM and non-linear SVM during validation and testing phases. Performance metric used during validation is accuracy. Performance metrics used during testing are accuracy and F1 score. Overall performance calculated across all 12 users using average, maximum, minimum, median, and standard deviation statistics.
ACC: Accuracy; LSVM: Linear Support Vector Machine; NLSVM: Non-Linear Support Vector Machine; VAL: Validation.}
\label{table:overall_results}
\vspace{-1em}
\end{table*}

\begin{figure}[t]
         \centering
         \fbox{\includegraphics[width=0.75\linewidth]{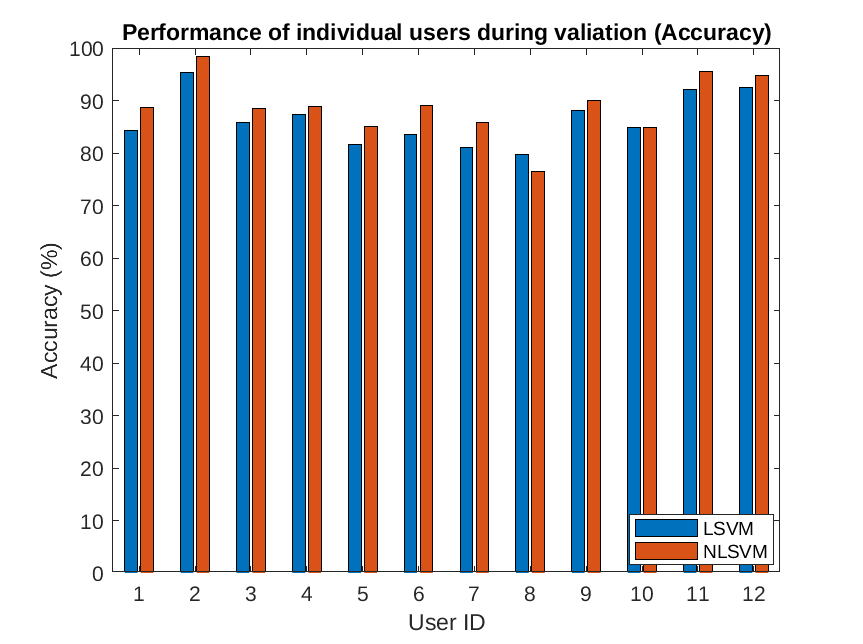}}
         \caption{\small Individual user's accuracy during validation using linear SVM (LSVM) and non-linear SVM (NLSVM).}
         \label{fig:val_acc}
\end{figure}

\subsection{Classification}
\label{sec:classification}

For user identification, we initially use kNN as the classifier but it produces around 75\% of accuracy.
Hence, we utilize both linear and non-linear SVMs for classification which enhanced the performance. We tune the classifier parameters using grid search. To tune kNN we use k = 4, 5, and 6 and distance metrics = Euclidean and Manhattan and perform grid search. 
The non-linear SVM has an RBF (Radial Basis Function) kernel and C = 0.1, 1, 10, and 100 and gamma = scale and auto are used for parameter tuning using grid search. The linear SVM does not have a gamma parameter and therefore the grid search is performed to tune C for which C values of 0.1, 1, 10, and 100 are chosen.

\section{RESULTS}
\label{sec:results}

\begin{figure}[h!]
    \centering
    \begin{subfigure}[t]{0.5\textwidth}
        \centering
        \fbox{\includegraphics[height=2.0in]{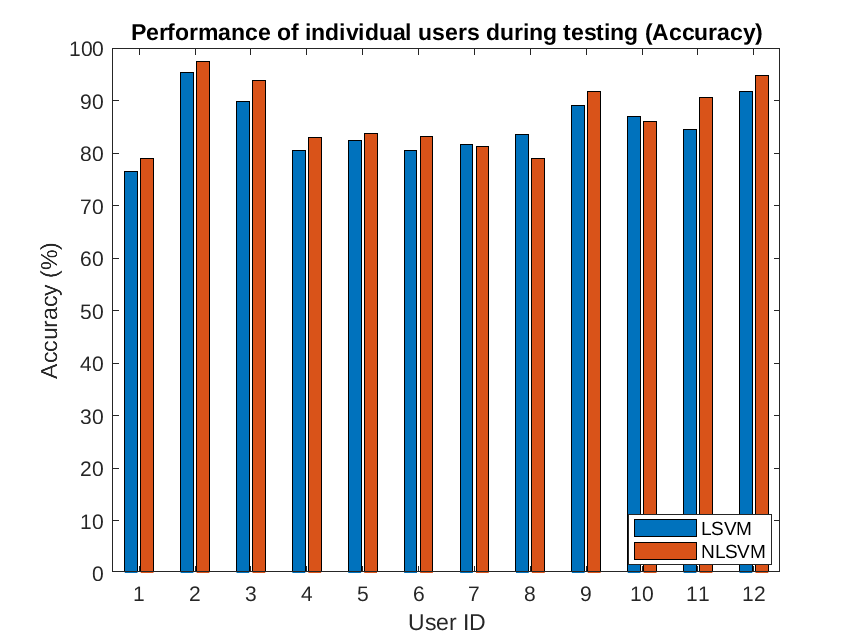}}
        \caption{}
        \label{fig:acc_sub}
    \end{subfigure}
    
\begin{subfigure}[t]{0.5\textwidth}
    \centering
    \fbox{\includegraphics[height=2.0in]{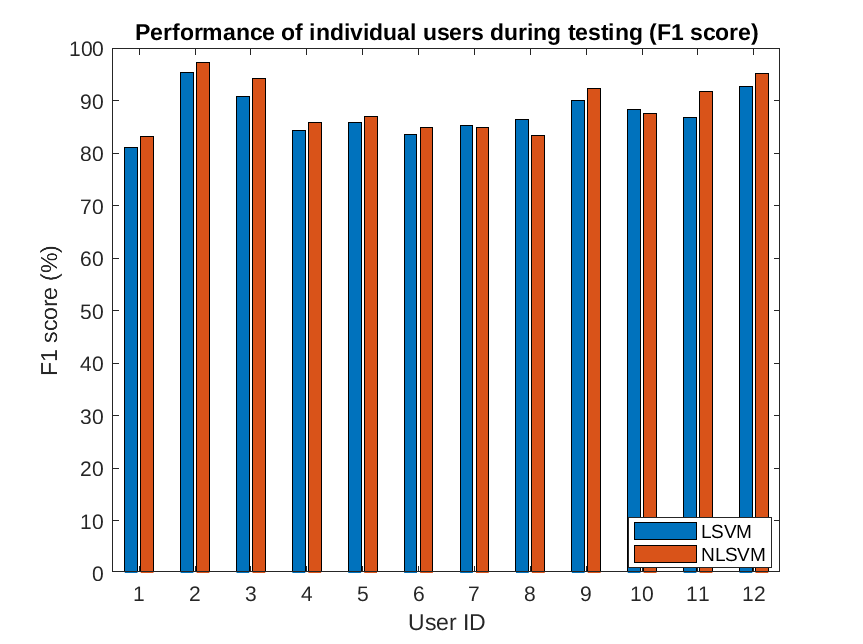}}
    \caption{}
    \label{fig:f1_sub}
    \end{subfigure}
    \caption{\small Individual user's performance using linear SVM (LSVM) and non-linear SVM (NLSVM). (a) Accuracy and (b) F1 score.}
    \label{fig:test}
\end{figure}

We identify each user as genuine against all other impostors both during validation and testing. 
Both linear and non-linear SVMs are utilized in this process.
The metric used for measuring performance during validation is accuracy.
We tune the classifier hyperparameters (as shown in Section~\ref{sec:classification}) through the validation step.
The individual performances of the 12 users during validation are shown in Figure~\ref{fig:val_acc}. From the plot, we can see that the non-linear SVM has performed better than the linear SVM although the performances of both classifiers are very close. From this, we understand that the data samples of genuine and impostors overlap and the non-linear SVM is precise enough to make a better boundary between the overlapped classes. 
We use both accuracy and F1 score to measure performance during testing. Figure~\ref{fig:test} shows individual performances of users using both linear and non-linear SVMs. From both testing accuracy (figure~\ref{fig:acc_sub}) and F1 scores (figure~\ref{fig:f1_sub}) we can understand that the non-linear SVM performs better than the linear SVM but here too both performances are very close. Table~\ref{table:overall_results} shows the overall performances of individual classifiers in the validation and testing phases. We have calculated the average, maximum, minimum, median, and standard deviation of performances across users to estimate the overall performance of the system. During testing, the non-linear SVM shows average accuracy and F1 score of 86.92\% and 88.94\% respectively which outperform the linear SVM performances (average accuracy of 85.18\% and average F1 score of 87.51\%). But in both cases of linear and non-linear SVMs, we have users whose individual performances are above 95\% in validation and testing. See Appendix~\ref{FifthAppendix} for the system's specifications in which all the experiments are ran.

\section{HYPOTHESIS TESTING}
\label{sec:hypo_test}

We perform hypothesis testing taking individual performances of users during testing using t-test. Before performing the t-test we verify the distribution of 12 performances (accuracy and F1 score) for both linear and non-linear SVMs are normally distributed using Kolmogorov-Smirnov (KS) test. In the KS tests, the distributions of linear SVM accuracies, non-linear SVM accuracies, linear SVM F1 scores, and non-linear SVM F1 scores show p values of  0.93, 0.65, 0.84, and 0.68 respectively. Each of the p value is greater than the significance level (0.05) so we accept the null (H0) hypothesis which states that the data does not differ significantly from that which is normally distributed. We, therefore, perform two t-tests one between the distributions of accuracies of linear and non-linear SVMs and the other between the distributions of F1 scores of linear and non-linear SVMs. The null (H0) and alternate (H1) hypotheses for the t-tests are stated below:

H0: \textit{The two models' (linear and non-linear SVMs) performances are not significantly different from each other.}

H1: \textit{The two models' (linear and non-linear SVMs) performances are significantly different from each other.}

In the t-test with accuracy distributions between linear and non-linear SVMs, the p-value is 0.48 and the other t-test between both SVMs using F1 score distribution is 0.44. Each of the p values is greater than 0.05 significance level. Hence we fail to reject (therefore we accept) the null (H0) hypothesis. This further ensures the close performances between the linear and non-linear SVMs.

\section{DISCUSSION AND CONCLUSION}
\label{sec:conclusion}

We examine the uniqueness and permanence characteristics of the EEG data from WAY\_EEG\_GAL~\cite{6-Luciw2014} dataset for user authentication. 
For each user, we select important features using ExtraTreeClassifier.
We have used both linear and non-linear SVMs for classification. The performances of the classifiers are very close to one another though non-linear SVM tends to perform better consistently. The hypothesis t-testing shows that the distribution of accuracies and F1 scores of the 12 users are not statistically significantly different and therefore the close performances of the classifiers are justified. The overall average accuracies of 85.18\% and 86.92\% are achieved using linear  and non-linear SVMs respectively. We observe average F1 scores of 87.51\%and 88.94\% for linear and non-linear SVMs respectively. Beyond overall performance, we also observe individuals showing high performances of 95.3\% accuracy (95.3\% F1 score) with  linear SVM and 97.4\% accuracy (97.3\% F1 score) with non-linear SVM.

Therefore, authentication through EEG data collected when users are performing grasp-and-lift mental tasks is feasible. In other words, we observe acceptable uniqueness and permanence trend in the WAY\_EEG\_GAL~\cite{6-Luciw2014} dataset. There are many other datasets which could have been chosen for this study, but the justification of selecting this dataset is to authenticate users while they are performing mental tasks of lifting objects for 2 hours where the weights and surfaces of the object change randomly. 

Through the observation made in this study, we pose the following open questions to the research community.
If the duration of the task performance per user is increased to more than 2 hours will the EEG based authentication accuracy be lowered than our obtained best? This question is toward this dataset and any other EEG datasets. In other words, can data collected over expanded time can pose a challenge to the permanence characteristics of EEG modality? 
If the user tasks performed during EEG data acquisition are not confined to rest state and monotonous mental tasks (like in WAY\_EEG\_GAL~\cite{6-Luciw2014}), and involve even more dynamic tasks, will the it impact the user authentication performance?

Given the above, it is always encouraged to examine 
further EEG datasets to test out their discriminability and permanence characteristics. Further testing across different datasets collected while users are performing wide ranges of tasks (in addition to examining rest state EEG datasets) can further strengthen the fact of whether it is feasible to deploy EEG based biometric systems in the public domain.

\bibliographystyle{ieee}

\bibliography{ms}

\appendix

\subsection{\textbf{EEG Electrode Labeling}}
\label{FirstAppendix}
For an EEG sensor device, each electrode placement location is identified by a letter (abbreviated representation) which denotes the brain lobe. The augmentation of the most common letters that denotes electrode placement locations are as follows. 
\begin{itemize}
   \item Fp: pre-frontal
   \item F:  frontal
   \item T:  temporal
   \item P:  parietal
   \item O:  occipital
   \item C:  central
   \item A/M:behind the outer ear  
\end{itemize}

The letter "Z" denotes electrode placement on the midline sagittal plane of the skull (FpZ, Fz, Cz, Oz).
Even numbered electrodes (2, 4, 6, 8) and odd numbered electrodes (1, 3, 5, 7) denote electrode placement on right side and left side of the skull respectively.

In high resolution EEG sensors, where more electrodes are used a combined nomenclature is used. For example, 
AF : located between Fp and F; FC : located between F and C; TP : located between T and P; 
PO : located between P and O~\cite{electrode_loc}.

The 32 electrode placement locations of WAY\_EEG\_GAL~\cite{6-Luciw2014}, in their data collection are: Fp1, Fp2, F7, F3, Fz, F4, F8, FC5, FC1, FC2, FC6, T7, C3, Cz, C4, T8, TP9, CP5, CP1, CP2, CP6, TP10, P7, P3, Pz, P4, P8, PO9, O1, Oz, O2, and PO10.

\subsection{\textbf{Visualization of raw EEG data in time domain}}
\label{SecondAppendix}

\begin{figure}[h]
         \centering
         \fbox{\includegraphics[width=0.9\linewidth]{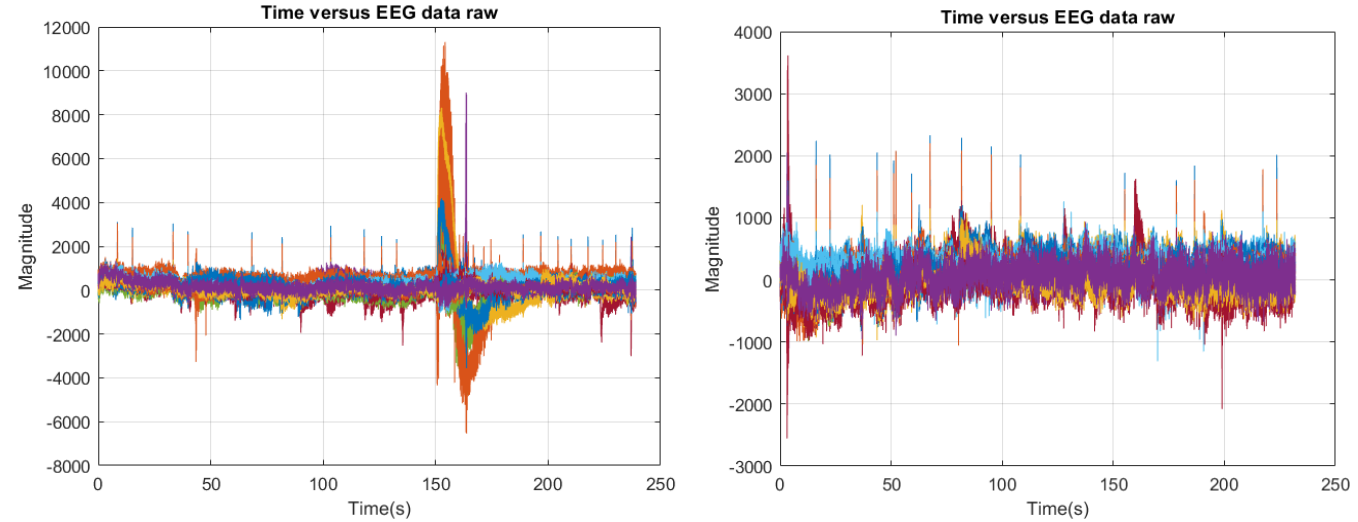}}
         \caption{\small Time domain EEG data of user 1 in session 1 (left) and session 9 (right).}
         \label{fig:P1_raw}
\end{figure}

\begin{figure}[h]
         \centering
         \fbox{\includegraphics[width=0.9\linewidth]{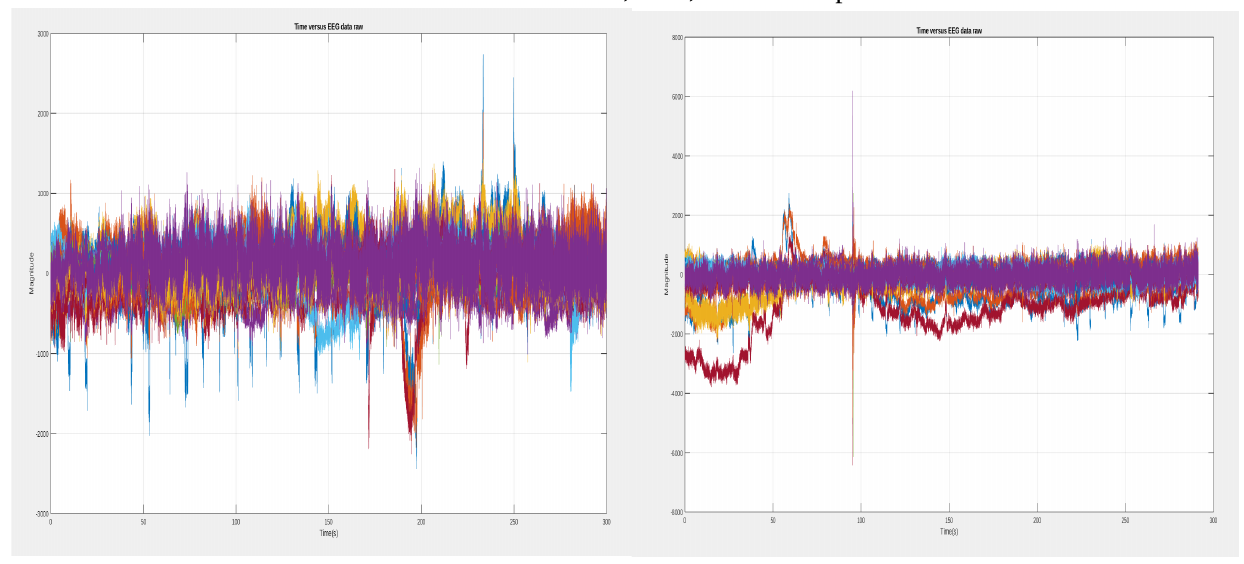}}
         \caption{\small Time domain EEG data of user 12 in session 1 (left) and session 9 (right).}
         \label{fig:P12_raw}
\end{figure}

Below figures show the raw EEG data in time domain of two sample users (1 and 12) for sessions 1 (the first session) and 9 (the last session).
The visualizations of raw EEG data in time domain for users 1 and 12 are in Figures~\ref{fig:P1_raw} and~\ref{fig:P12_raw}.

\subsection{\textbf{Visualization of features}}
\label{ThirdAppendix}

\begin{figure}[h]
         \centering
         \fbox{\includegraphics[width=0.9\linewidth]{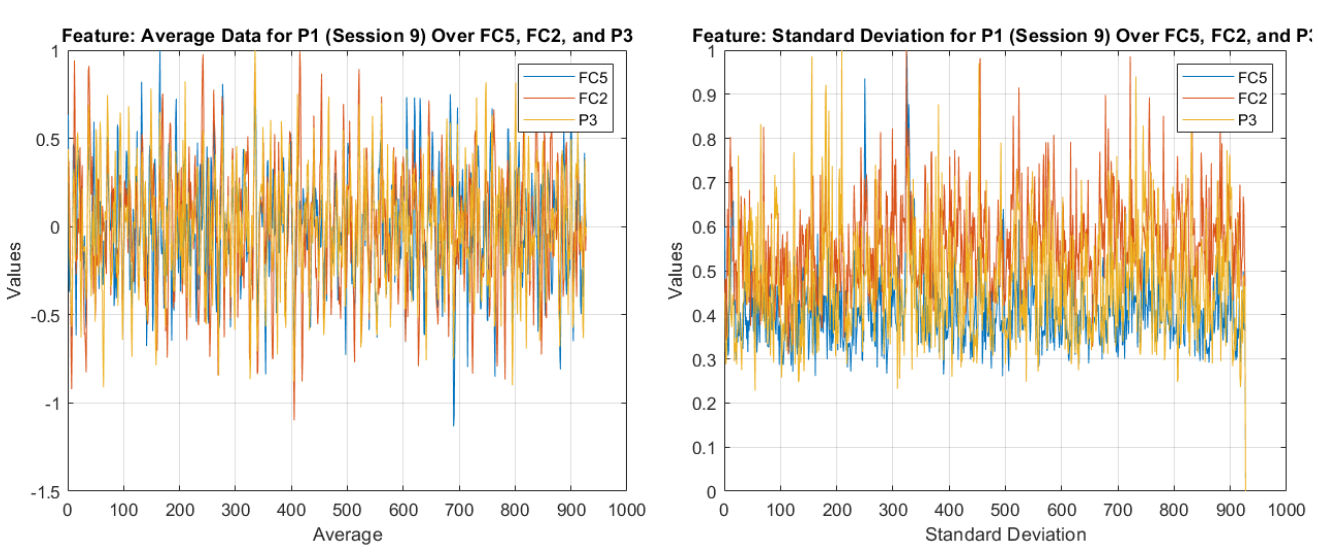}}
         \caption{\small The Average (left) and Standard Deviation (right) features of User-1 in Session 9.}
         \label{fig:P1_avg_std}
\end{figure}

\begin{figure}[h]
         \centering
         \fbox{\includegraphics[width=0.9\linewidth]{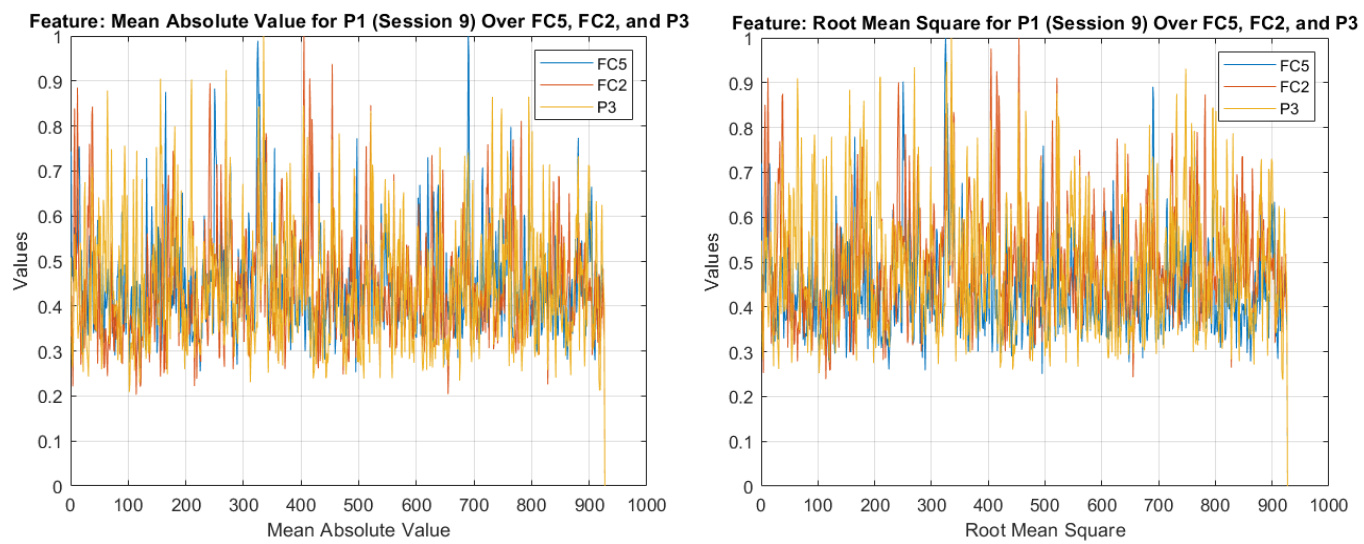}}
         \caption{\small The Mean Absolute Value (left) and Root Mean Square (right) features of User-1 in Session 9.}
         \label{fig:P1_mav_rms}
\end{figure}

\begin{figure}[h]
         \centering
         \fbox{\includegraphics[width=0.9\linewidth]{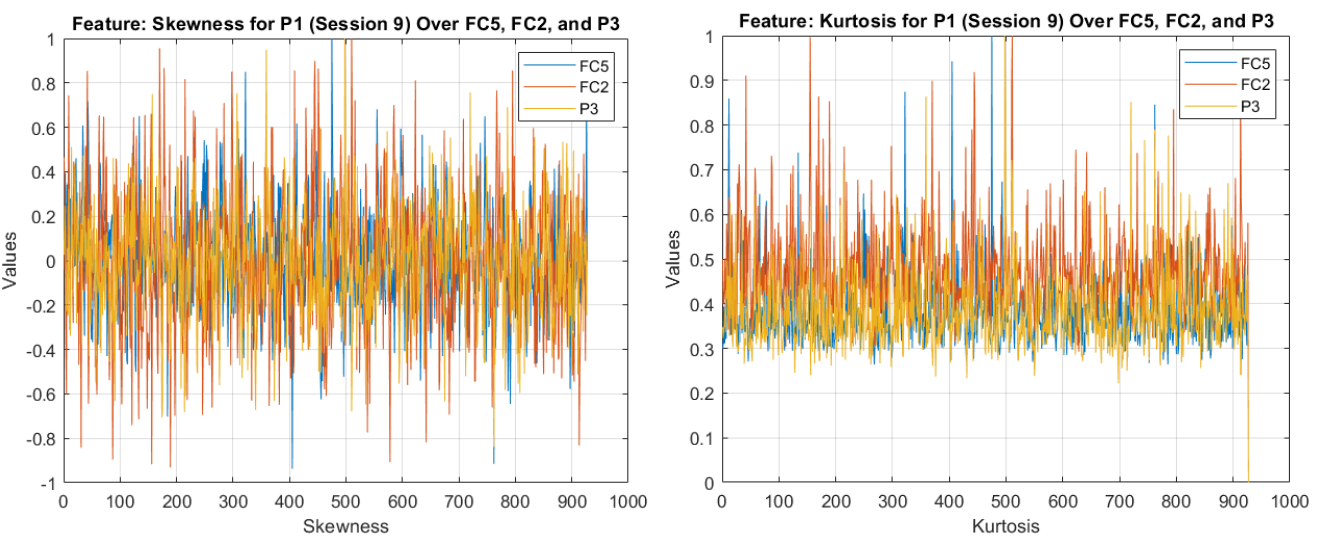}}
         \caption{\small The Skewness (left) and Kurtosis (right) features of User-1 in Session 9.}
         \label{fig:P1_skw_kur}
\end{figure}

\begin{figure}[h]
         \centering
         \fbox{\includegraphics[width=0.9\linewidth]{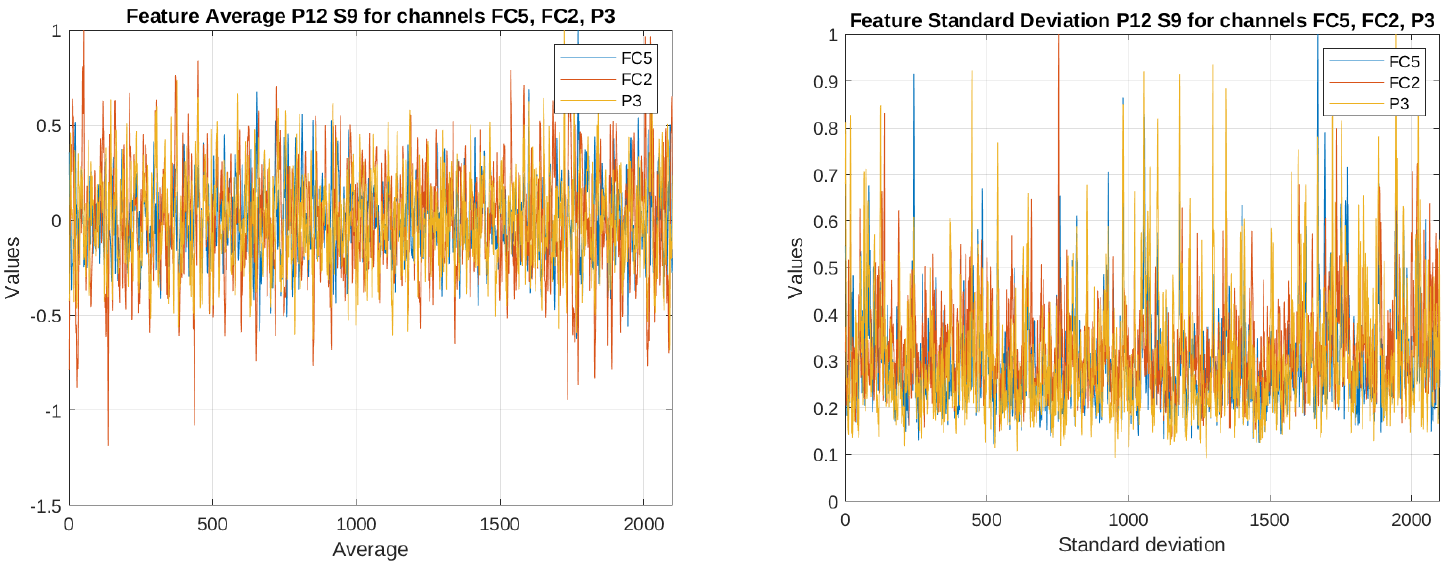}}
         \caption{\small The Average (left) and Standard Deviation (right) features of User-12 in Session 9.}
         \label{fig:P12_avg_std}
\end{figure}

\begin{figure}[h]
         \centering
         \fbox{\includegraphics[width=0.9\linewidth]{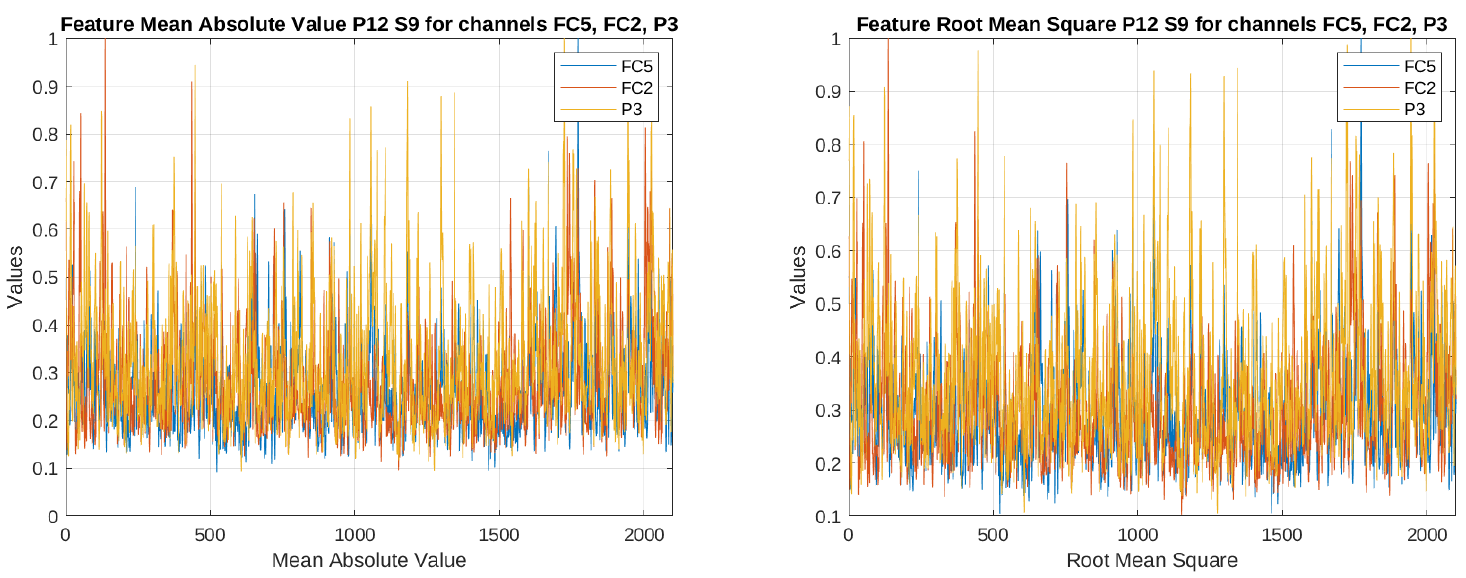}}
         \caption{\small The Mean Absolute Value (left) and Root Mean Square (right) features of User-12 in Session 9.}
         \label{fig:P12_mav_rms}
\end{figure}

\begin{figure}[h]
         \centering
         \fbox{\includegraphics[width=0.9\linewidth]{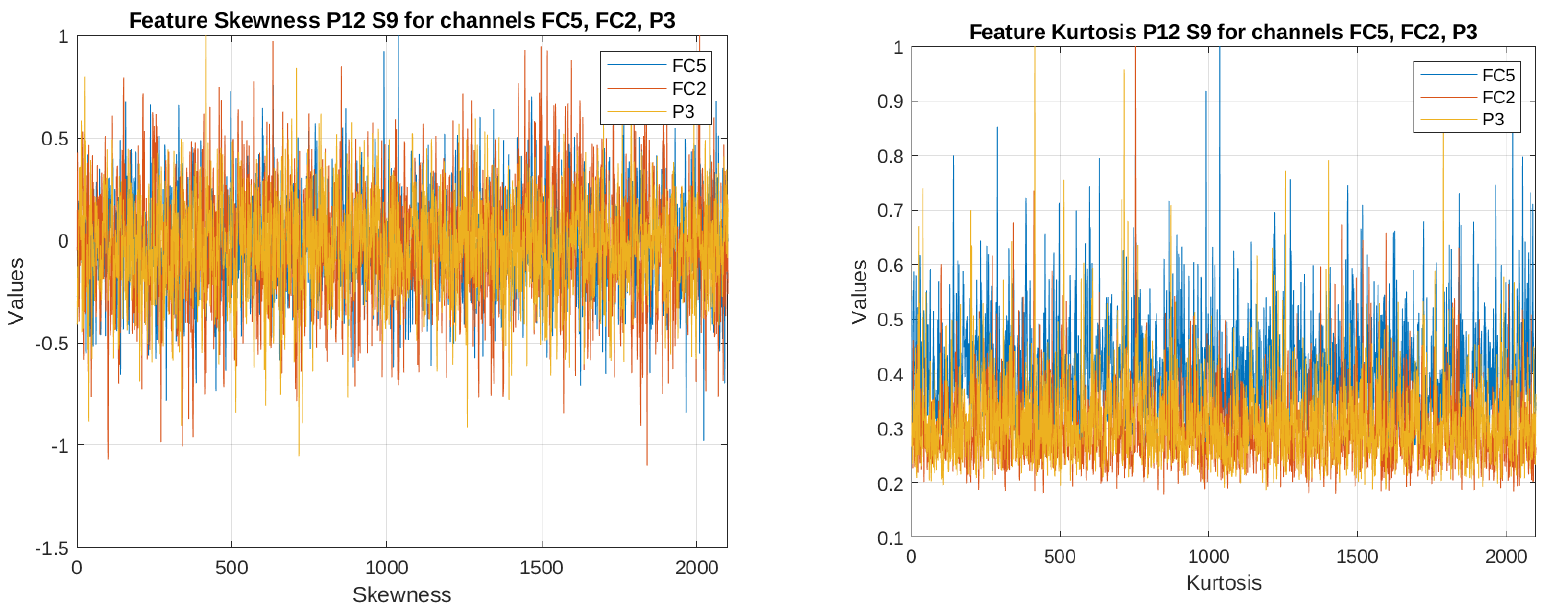}}
         \caption{\small The Skewness (left) and Kurtosis (right) features of User-12 in Session 9.}
         \label{fig:P12_skw_kur}
\end{figure}

Below figures show the features extracted from WAY\_EEG\_GAL~\cite{6-Luciw2014} data in session 9 (the last session) of two sample users (users 1 and 12).
Only plots from channels FC5, FC2, and P3, since these channels gets most active while lifting objects using precision grip~\cite{kinoshita2000functional}.
Figures~\ref{fig:P1_avg_std},~\ref{fig:P1_mav_rms}, and~\ref{fig:P1_skw_kur} are features for user-1 in session-9.
Figures~\ref{fig:P12_avg_std},~\ref{fig:P12_mav_rms}, and~\ref{fig:P12_skw_kur} are features for user-12 in session-9.

\subsection{\textbf{Visualization of Sample Feature Importance}}
\label{FourthAppendix}

\begin{figure}[h]
         \centering
         \fbox{\includegraphics[width=0.55\linewidth]{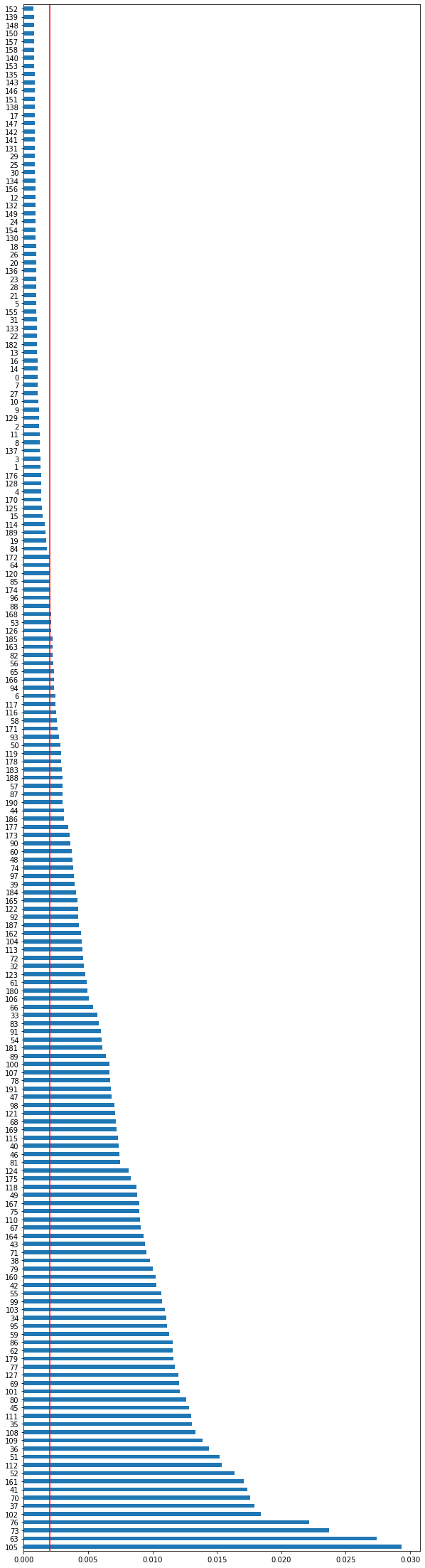}}
         \caption{\small Sample Feature importance based on tuned threshold.}
         \label{fig:feat_thresh}
\end{figure}

Figure~\ref{fig:feat_thresh} shows the selection of user specific features  through ExtraTreeClassifier feature importance algorithm. The sample user here is user-12. The threshold selection is pre-tuned per user which is shown as the red vertical line in the image. 

\subsection{\textbf{System Specifications}}
\label{FifthAppendix}

The system configuration for the local server used to run experiments are as follows. 
The system is an HP Z620 workstation with 92 GB RAM and 2 Intel Xeon E5-2670 2.6 GHz 16 core processors. 
Therefore, there are 32 cores. 
It also has a small-scale GPU namely, Nvidia Quadro 600. 
Although there is no use of GPUs in this experimental setup.
The GPU is used to operate the system's display.
It has Kubuntu 18.04.5 operating system. 
There are a few more additional setup done to the system for storage purposes. 
An SSD of 500 GB capacity is added for the operating system. 
Additionally, an HDD of 2TB capacity (WD Blue) is added for backup
and/or bulk storage.

\end{document}